%% file: main.tex
\documentclass[11pt,a4paper]{article}
\usepackage[hyperref]{emnlp2020}
\usepackage{times}
\usepackage{latexsym}

\aclfinalcopy % Uncomment this line for the final submission
\def\aclpaperid{*}
\usepackage{url}
\usepackage{algorithm}
\usepackage{algorithmic}
\usepackage{amsmath}
\usepackage{multirow}
\usepackage{tabu}
\usepackage{booktabs}
\usepackage{graphicx}
\usepackage{paralist}
\usepackage{enumitem}
\usepackage{adjustbox}
\usepackage{comment}

\title{Robustness to Modification with Shared Words\\ in Paraphrase Identification}

\author{Zhouxing Shi\\
  DCST, Tsinghua University, \\
  Beijing 100084, China \\  
  \texttt{zhouxingshichn@gmail.com} \\\And
  Minlie Huang\thanks{\quad Corresponding author}  \\
  DCST, THUAI, SKLits, BNRist \\
  Tsinghua University, \\
  Beijing 100084, China \\
  \texttt{aihuang@tsinghua.edu.cn} }

\date{}

\begin{document}

\maketitle

\begin{abstract}
Revealing the robustness issues of natural language processing models and improving their robustness is important to their performance under difficult situations. In this paper, we study the robustness of paraphrase identification models from a new perspective -- via modification with shared words, and we show that the models have significant robustness issues when facing such modifications. To modify an example consisting of a sentence pair, we either replace some words shared by both sentences or introduce new shared words. We aim to construct a valid new example such that a target model makes a wrong prediction. To find a modification solution, we use beam search constrained by heuristic rules, and we leverage a BERT masked language model for generating substitution words compatible with the context. Experiments show that the performance of the target models has a dramatic drop on the modified examples, thereby revealing the robustness issue. We also show that adversarial training can mitigate this issue. 
\end{abstract}

\input{introduction.tex}

\input{related_work.tex}

\input{method.tex}

\input{experiments.tex}

\input{conclusion.tex}

\section*{Acknowledgments}
This work was jointly supported by the NSFC projects (Key project with No. 61936010 and regular project with No. 61876096), and the Guoqiang Institute of Tsinghua University, with Grant No. 2019GQG1.
We thank THUNUS NExT Joint-Lab for the support.

\bibliography{main}
\bibliographystyle{acl_natbib}

\appendix
\clearpage
\newpage
\input{appendix.tex}

\end{document}

%% file: introduction.tex
\section{Introduction}

Paraphrase identification is to determine whether a pair of sentences have the same meaning~\citep{socher2011dynamic},
with many applications such as duplicate question matching on social media~\citep{quora} and plagiarism detection \citep{clough2000plagiarism}.
It can be viewed as a sentence matching problem, and many neural models have achieved great performance on benchmark datasets~\citep{wang2017bilateral,gong2017natural,devlin2018bert}.

Despite this progress, there is not much work on the \emph{robustness} of paraphrase identification models, while natural language processing (NLP) models on other tasks have been shown to be vulnerable and lack of robustness.
In previous works for the robustness of NLP models, constructing semantic-preserving perturbations to input sentences while making the model prediction significantly change appears to be a popular way, in tasks such as text classification and natural language inference~\citep{alzantot2018generating,jin2019bert}.
However, on specific tasks, it is possible to design modification that is \emph{not} necessarily semantic-preserving, which can further reveal more robustness issues.
For instance, on reading comprehension, \citet{jia2017adversarial} conducted modification by inserting distracting sentences to the input paragraphs.
Such findings can be important for investigating and resolving the weakness of NLP models.

On paraphrase identification, to the best of our knowledge, the only previous work is PAWS~\citep{zhang2019paws} with a cross-lingual version~\citep{yang2019paws}, which found that models often make false positive predictions when words in  the two sentences only differ by word order.
However, this approach is for negative examples only, and for positive examples, they  used back-translation to still generate semantically similar sentences.
Moreover, it was unknown whether models still easily make false positive predictions when the word overlap between the two sentences is much smaller than 100\%.

\begin{figure}[ht]
  \centering
  \resizebox{.48\textwidth}{!}{
  \begin{tabular}{ll}
    \toprule
    (P) & What is ultimate \textbf{purpose} of \textbf{life}?\\
    (Q) & What is the \textbf{purpose} of \textbf{life} , if not money?\\
    (P') & What is ultimate \textbf{measure} of \textbf{value}?\\
    (Q') & What is the \textbf{measure} of \textbf{value} , if not money?\\
    Label & \textit{Positive}\\
    Output & \textit{Positive} (99.4\%) $\rightarrow$ \textit{Negative} (85.2\%)\\
    \midrule
    (P) & How can I get my \textbf{Gmail} \textbf{account} back ?\\
    (Q) & What is the best \textbf{school} \textbf{management} software ? \\
    (P') & How can I get my \textbf{credit score} back ?\\
    (Q') & What is the best \textbf{credit score} software ?\\
    Label & \textit{Negative}\\
    Output & \textit{Negative} (100.0\%) $\rightarrow$ \textit{Positive} (68.3\%)\\
    \bottomrule
  \end{tabular}
  }
  \caption{
  Examples with labels \textit{positive} and \textit{negative} respectively, originally from Quora Question Pairs (QQP) \citep{quora}.
  ``(P)'' and ``(Q)'' are original sentences while ``(P')'' and ``(Q')'' are modified.
  Modified words are highlighted in bold.
  ``Output'' indicates the change of output labels by BERT \citep{devlin2018bert}, where the percentage numbers are confidence scores.
  }
  \label{intro_example}
\end{figure}
%\vspace{-0.2in}

In this paper, we propose an algorithm for studying the robustness of paraphrase identification models from a new perspective -- via modifications with \textbf{shared words (words that are shared by both sentences)}.
For positive examples, i.e., the two sentences are paraphrases, we aim to see whether models can still make correct predictions when some shared words are replaced.
Each pair of selected shared words are replaced  with a new word, and the new example tends to remain positive.
As the first example in Figure~\ref{intro_example} shows, by replacing ``purpose'' and ``life'' with ``measure'' and ``value'' respectively, the sentences change from asking about ``purpose of life'' to ``measure of value'' and remain paraphrases, but the target model makes a wrong prediction.
This indicates that the target model has a weakness in generalizing from ``purpose of life'' to ``measure of value''.
On the other hand, for negative examples, we replace some words and introduce new shared words to the two sentences while trying to keep the new example negative.
As the second example in Figure~\ref{intro_example} shows, with new shared words ``credit'' and ``score'' introduced, the new example remains negative but the target model makes a false positive prediction.
This reveals that the target model can be distracted by the shared words while ignoring the difference in the unmodified parts.
The unmodified parts of the two sentences have a low word overlap to reveal such a weakness.
In contrast, examples in PAWS had exactly the same bag of words and are not capable for this investigation.

In our word replacement, to preserve the label and language quality, we impose heuristic constraints on replaceable positions.
Furthermore, we apply a BERT masked language model~\citep{devlin2018bert} to generate substitution words compatible with the context. 
We use beam search to find a word replacement solution that approximately maximizes the loss of the target model and thereby tends to make the model fail.

We summarize our contributions below:

\begin{itemize}[labelwidth=!, labelindent=0pt, itemsep=0pt]
    \item We study the robustness of paraphrase identification models via modification with shared words. Experiments show that models have a severe performance drop on our modified examples, which reveals a robustness issue.
    \item We propose a novel and concise method that leverages the BERT masked language model for generating substitution words compatible with the context.
    \item We show that adversarial training with our generated examples can mitigate the robustness issue.
    \item Compared to previous works, our perspective is new: 1) Our modification is not limited to be semantic-preserving; and 2) Our negative examples have much lower word overlap between two sentences, compared to PAWS.
\end{itemize}

%% file: related_work.tex
\section{Related Work}

\subsection{Paraphrase Identification Models}

There exist many neural models for sentence matching and paraphrase identification.
Some works applied a classifier on independently-encoded embeddings of  two sentences~\citep{snli:emnlp2015,yang2015wikiqa,conneau2017supervised}, and some others made strong interactions between the two sentences by jointly encoding and matching them~\citep{wang2017bilateral,duan2018attention,kim2018semantic} or hierarchically extracting features from their interaction space~\citep{hu2014convolutional,pang2016text,gong2017natural}.
Notably, 
BERT pre-trained on large-scale corpora achieved even better results~\citep{devlin2018bert}.

\subsection{Robustness of NLP Models}
\label{difference}

On the robustness of NLP models, 
many previous works constructed semantic-preserving perturbations to input sentences~\citep{alzantot2018generating,iyyer2018adversarial,ribeiro2018semantically,hsieh2019robustness,jin2019bert,ren2019generating}.
However, NLP models for some tasks have robustness issues not only when facing semantic-preserving perturbations.
In reading comprehension, \citet{jia2017adversarial} studied the robustness issue when a distractor sentence is added to the paragraph.
In natural language inference, \citet{minervini2018adversarially} considered logical rules of sentence relations, and \citet{glockner2018breaking} used single word replacement with lexical knowledge.
Thus methods for general NLP tasks alone are insufficient for studying the robustness of specific tasks.
In particular, for paraphrase identification, the only prior work is PAWS~\citep{zhang2019paws,yang2019paws} which used word swapping, but this method is for negative examples only and each constructed pair of sentences have exactly the same words.

%% file: method.tex
\section{Methodology}

\subsection{Algorithm Framework}

\label{sec:framework}

Paraphrase identification can be formulated as follows:
given two sentences $P=p_1p_2\cdots p_n$ and $Q=q_1q_2\cdots q_m$, the goal is to predict whether $P$ and $Q$ are paraphrases.
The model outputs a score $[Z (P, Q)]_{\hat{y}}$ for each class $\hat{y}\in \mathcal{Y} = \{positive, negative \}$, where \emph{positive} means $P$ and $Q$ are paraphrases and vice versa.

We first sample an original example from the dataset and then conduct modification.
We take multiple steps for modification until the model fails or the step number limit is reached.
In each step, we replace a word pair with a shared word,
and we evaluate different options according to the model loss they induce.
We use beam search to find approximately optimal options.
The modified example evaluated as the best option is finally returned.

In the remainder of this section, we introduce what modification options are considered available to our algorithm in Sec. \ref{sec:options} and how to find optimal modification solutions in Sec. \ref{sec:find_sol}.

\subsection{Modification Options}
\label{sec:options}

\paragraph{Original Example Sampling}
\label{original_example_sampling}

To sample an original example from the dataset, for a positive example, we directly sample a \emph{positive} example from the original data, namely, $(P, Q, positive)$; and for a negative example, we sample two different sentence pairs $(P_1, Q_1)$ and $(P_2, Q_2)$, and we then form a negative example $(P_1, Q_2, negative)$.

\paragraph{Replaceable Position Pairs}

\begin{figure}[ht]
  \centering
  \resizebox{.48\textwidth}{!}{
    \includegraphics[width=\textwidth]{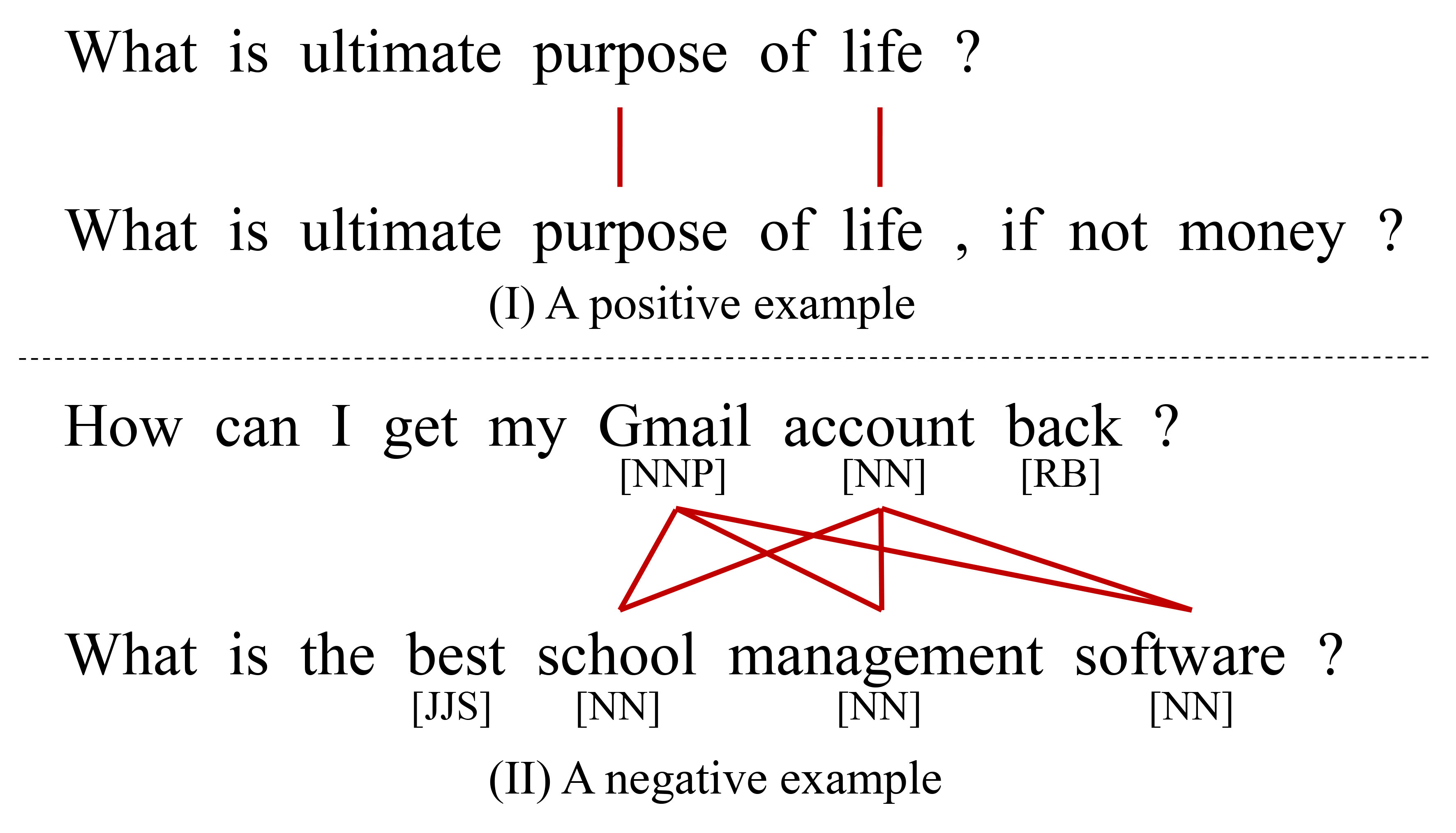}  
  }
  \caption{
  Examples of identifying replaceable position pairs that are linked with red lines.
  In the negative example, POS tags of non-stopwords are also shown.
  }
  \label{replace}
\end{figure}

\noindent For a sentence pair under modification, we impose heuristic rules on replaceable position pairs.
First, we do not replace stopwords.
Besides, for a positive example, we require each replaceable word pair to be shared words, while for a negative example, we only require them to be both nouns, both verbs, or both adjectives, according to Part-of-Speech (POS) tags obtained using Natural Language Toolkit (NLTK) \citep{bird2009natural}.
Two examples are shown in Figure~\ref{replace}. 
For the first example (positive), only shared words ``purpose'' and ``life'' can be replaced,
and the two modified sentences are likely to talk about another \emph{same} thing, e.g. changing from ``purpose of life'' to ``measure of value'', and thereby the new example tends to remain \emph{positive}.
As for the second example (negative), nouns ``Gmail'', ``account'', ``school'', ``management'' and ``software'' can be replaced.
Consequently, the modified sentences are based on templates ``How can I get $\cdots$ back ? '' and ``What is the best $\cdots$ ?'', and the pair tends to remain negative even if the template is filled by shared words.
In this way, the labels can usually be preserved.

\paragraph{Substitution Words}

We use a pre-trained BERT masked language model \citep{devlin2018bert} to generate substitution words compatible with the context, for each replaceable position pair.
Specifically, to replace word $p_i$ and $q_j$ from the two sentences respectively with some shared word $w$, we compute a joint probability distribution
\begin{align*}
    &\mathcal{P}(w| p_{1:i-1}, p_{i+1:n}, q_{1:j-1}, q_{j+1:m})\\
    = &\mathcal{P} (w | p_{1:i-1}, p_{i+1:n}) \cdot 
    \mathcal{P} (w | q_{1:j-1}, q_{j+1:m}),
\end{align*}
where $s_{i:j}$ denotes the subsequence starting from $i$ to $j$.
{\small$\mathcal{P} (w | p_{1:i-1}, p_{i+1:n})$} and {\small$\mathcal{P} (w | q_{1:j-1}, q_{j+1:m})$} are obtained from the language model by masking $p_i$ and $q_j$ respectively.
We rank all the words in the vocabulary of the model and choose top $K$ words with largest probabilities, as the candidate substitution words for the position pair.

This method of generating substitution words enables us to find out possible substitution words and also verify their compatibility with the context simultaneously, compared to previous methods that have these two separated~ \citep{alzantot2018generating,jin2019bert} -- they first constructed a candidate substitution word list from synonyms, and using each substitution word respectively, they then checked the language quality or semantic similarity constraints of the new sentence.
Moreover, some recent works~\citep{li2020bert,garg2020bae} that appeared later than our preprint have shown that using a masked language model for substituting words can outperform state-of-the-art methods in generating adversarial examples on text classification and natural language inference tasks. 

\begin{table*}[ht]
  \centering
  \caption{Accuracies (\%) of target models:
  ``Original full'' indicates the full original test set, ``original sampled'' indicates original examples sampled from the test set (see Sec. \ref{original_example_sampling}), and ``modified'' indicates examples modified by our algorithm.
  ``Pos'' and ``neg'' indicate results on positive examples and negative examples respectively.
The ``training'' column indicates whether the models are normally trained or adversarial trained (see Sec. \ref{sec:exp_adv_train}).
  }
  \adjustbox{max width=.9\textwidth}{
  \begin{tabular}{c|l|c|ccc|ccc|ccc}
    \toprule[1pt]
    \multirow{2}{*}{Dataset}&
    \multirow{2}{*}{Target Model}&
    \multirow{2}{*}{Training}&
    \multicolumn{3}{c|}{Original full} &
    \multicolumn{3}{c|}{Original sampled} &
    \multicolumn{3}{c}{Modified}\\
    & & & Pos & Neg & All & Pos & Neg & All & Pos & Neg & All \\
    \hline
    \multirow{6}{*}{QQP} & 
    BiMPM & \multirow{3}{*}{Normal} & 88.5 & 87.8 & 88.1 & 88.0 & 99.4 & 93.7 & 14.4 & 7.8 & 11.1\\ 
    & DIIN &  & 91.5 & 85.9 & 88.7 & 89.6 & 99.6 & 94.6 & 31.0 & 8.2 & 19.6\\
    & BERT &  & 90.7 & 91.3 & 91.0 & 89.0 & 99.6 & 94.3 & 33.4 & 14.8 & 24.1 \\ 
    \cline{2-12}
    & BiMPM & \multirow{3}{*}{Adversarial} & 89.6 & 88.0 & 88.9 & 89.4 & 99.8 & 94.6 & 15.0 & 27.8 & 21.4\\ 
    & DIIN & & 82.1 & 91.7 & 86.9 & 81.2 & 99.8 & 90.5 & 35.0 & 72.2 & 53.6\\ 
    & BERT & & 87.6 & 92.5 & 90.1 & 86.8 & 99.8 & 93.3 & 53.0 & 79.0 & 66.0\\ 
    \hline
    \multirow{6}{*}{MRPC} &  
    BiMPM &  \multirow{3}{*}{Normal} & 90.2 & 40.0 & 73.4 & 87.2 & 97.4 & 92.3 & 3.2 & 0.2 & 1.7\\ 
    & DIIN  & & 89.9 & 49.5 & 76.3 & 90.4 & 100.0 & 95.2 & 48.2 & 0.4 & 24.3 \\
    & BERT  & & 93.2 & 66.4 & 84.2 & 94.0 & 100.0 & 97.0 & 45.6 & 2.0 & 23.8 \\
    \cline{2-12}
    & BiMPM &  \multirow{3}{*}{Adversarial} & 96.8 & 26.3 & 73.2 & 95.6 & 100.0 & 97.8 & 73.2 & 0.6 & 36.9\\ 
    & DIIN & & 85.8 & 58.0 & 76.5 & 82.8 & 100.0 & 91.4 & 59.8 & 67.6 & 63.7\\
    & BERT & & 95.3 & 55.2 & 81.9 & 95.0 & 100.0 & 97.5 & 81.0 & 93.0 & 87.0\\
    \bottomrule[1pt]
  \end{tabular}
  }
  \label{main_results}
\end{table*}

\subsection{Finding Modification Solutions}

\label{sec:find_sol}

We then use beam search with beam size $B$ to find a modification solution in multiple steps.
At step $t$, we have two stages to determine the replaced positions and the substitution words respectively, based on a two-stage framework~\citep{yang2018greedy}.

First, for replaced positions, we enumerate all replaceable position pairs and  replace words on each pair of positions with a special token \textit{[PAD]} respectively.
We then query the model with these new examples and take top $B$ examples that minimizes the output score of the gold label.
Next, we enumerate all words in the candidate substitution word set of positions with \textit{[PAD]} and replace \textit{[PAD]} with each candidate substitution word respectively.
We again query the model with the examples after each possible replacement, and we take top $B$ examples similarly as in the first stage.
For the topmost example, if the label predicted by the model is already incorrect, we finish the modification process.
Otherwise, we take more steps until the model fails or the step number limit $S$ is reached.

%% file: experiments.tex
\section{Experiments}

\subsection{Datasets and Target Models}

We adopt two datasets. The Quora Question Pairs, \textbf{QQP}~\citep{quora}, consists of 384,348/10,000/10,000 question pairs in the training/development/test set as we follow the partition in \citet{wang2017bilateral}. And the Microsoft Research Paraphrase Corpus, \textbf{MRPC}~\citep{dolan2005automatically},  consists of sentence pairs from news with 4,076/1,725 pairs in the training/test set. 
Each sentence pair is annotated with a label indicating whether the two sentences are paraphrases or not (\emph{positive} or \emph{negative}).

We study three typical models for paraphrase identification.
\textbf{BiMPM}~\citep{wang2017bilateral} matches two sentences from multiple perspectives using BiLSTM layers.
\textbf{DIIN}~\citep{gong2017natural} adopts DenseNet \citep{huang2017densely} to extract interaction features.
\textbf{BERT}~\citep{devlin2018bert} is a pre-trained encoder fine-tuned on this task with a classifier applied on encoded representations. 
These models are representative in terms of backbone neural architectures: BiMPM is based on recurrent neural networks, DIIN on convolutional neural networks, and BERT on Transformers.

\subsection{Performance on Modified Examples}

We train each model on the original training set and then try to construct modification that makes the models fail.
For each dataset, we sample 1,000 original examples with balanced labels from the test set, and we modify them for each model.
We evaluate the accuracies of the models on our modified examples.
Table \ref{main_results} shows the results.
We focus on rows with ``normal'' for column ``training'' in this section. 
The models have high overall accuracies on the original data, 
but their performance drops dramatically on our modified examples (e.g., the overall accuracy of BERT on QQP drops from 94.3\% to 24.1\%).
This demonstrates that the models indeed have the robustness issue we aim to reveal.
Some examples are provided in Appendix \ref{appendix:cases}.

\subsection{Adversarial Training}

\label{sec:exp_adv_train}

To improve the model robustness, we further fine-tune the models using adversarial training.
A training batch consists of original examples and modified examples from the training data, where modified examples account for around 10\% in a batch.
The proportion of modified examples is directly chosen to demonstrate the effectiveness of adversarial training while preventing the model from overfitting on modified examples.
During training, we modify examples with the current model as the target and update the model parameters iteratively.
The beam size for generation is set to 1 to reduce the computational cost.
We evaluate the adversarially trained models as shown in Table \ref{main_results} (rows with ``adversarial'' for column ``training'').
The performance on modified examples of all the models raises significantly (e.g. the overall accuracy of BERT on modified examples raises from 24.1\% to 66.0\% for QQP and from 23.8\% to 87.0\% for MRPC).
This demonstrates that adversarial training with our modified examples can significantly improve the robustness, yet without remarkably hurting the performance on original data.
An improvement on the \emph{original} data is not expected since they cannot reflect robustness and it is even common to see a small drop in previous works~\citep{jia2017adversarial,iyyer2018adversarial,ribeiro2018semantically}.

\subsection{Manual Evaluation}

\begin{table}[ht]
 \centering
 \caption{Manual annotation results on original examples and modified examples respectively, including accuracies and grammaticality ratings.}
 \adjustbox{max width=.48\textwidth}{
 \begin{tabular}{c|l|cc}
    \hline
    Dataset & Metric & Original & Modified\\
    \hline
    \multirow{4}{*}{QQP} 
    & Accuracy - Pos & 86\% & 70\%\\
    & Accuracy - Neg & 98\% & 88\%\\
    & Accuracy - All & 92\% & 79\%\\
    \cline{2-4}
    & Grammaticality & 2.48 & 2.15\\
    \hline
    \multirow{4}{*}{MRPC} 
    & Accuracy - Pos & 90\% & 94\%\\
    & Accuracy - Neg & 100\% & 82\%\\
    & Accuracy - All & 95\% & 88\%\\
    \cline{2-4}
    & Grammaticality & 2.40 & 2.19\\
    \hline
 \end{tabular}
 }
 \label{table:human_evaluation}
 \end{table}

\noindent We also manually  verify the quality of the modified examples in terms of the label correctness and grammaticality.
For each dataset, using BERT as the target, we randomly sample 100 modified examples with balanced labels such that the model makes wrong predictions, and we present each of them to three workers on Amazon Mechanical Turk.
We ask the workers to label the examples and also rate the grammaticality of the sentences with a scale of 1/2/3.
We integrate annotations from different workers with majority voting for labels and averaging for grammaticality. 
Results are shown in Table~\ref{table:human_evaluation}.
We observe that the workers achieve acceptable accuracies on our modified examples (79\% on QQP and 88\% on MRPC), while their performance on original examples is not perfect either (92\% on QQP and 95\% on MRPC). 
The grammaticality drop between original examples and modified examples is also satisfactory (from 2.48 to 2.15 on QQP and from 2.40 to 2.19 on MRPC).
These results suggest that the labels and  grammaticality of the modified examples can be preserved with an acceptable quality.

%% file: conclusion.tex
\section{Conclusion}

In this paper, we present a novel algorithm to study the robustness of paraphrase identification models.
We show that the target models have a robustness issue when facing modification with shared words.
Such modification is substantially different from those in previous works -- the modification is not semantic-preserving and each pair of modified sentences generally have a much lower word overlap, and thereby it reveals a new robustness issue.
We also show that model robustness can be improved using adversarial training with our modified examples.

%% file: appendix.tex
\section{Implementation Details}

We adopt open source codes for BiMPM\footnote{\url{https://github.com/zhiguowang/BiMPM}}, DIIN\footnote{\url{https://github.com/YichenGong/Densely-Interactive-Inference-Network}} and BERT ($\rm BERT_{base}$ is used)\footnote{\url{https://github.com/huggingface/transformers}}, and the datasets are downloaded from the internet for both QQP\footnote{\url{https://www.quora.com/q/quoradata/First-Quora-Dataset-Release-Question-Pairs}} and MRPC\footnote{\url{https://www.microsoft.com/en-us/download/details.aspx?id=52398}}.
There are 1.4, 42.8, and 109.5 million parameters in BiMPM, DIIN and BERT respectively.

For QQP, the step number limit of modification, $S$, is set to 5; the number of candidate substitution words suggested by the language model, $K$, and the beam size $B$ are both set to 25.
$S$, $K$ and $B$ are doubled for MRPC where sentences are generally longer.

We conduct the experiments on an NVIDIA TITAN X GPU.
On QQP, the average time cost per example is around 4.7s for positive examples and 7.5s for negative examples.
On MRPC, it is around 44.9s for positive examples and 61.6s for negative examples.

\section{Examples of Our Modifications}

\label{appendix:cases}

\begin{table}[!ht]
  \centering
  \caption{
  Modified examples for BERT as the target model on QQP.
  ``(P)'' and ``(Q)'' indicate original sentences, and ``(P')'' and ``(Q')'' indicate modified sentences.
  Modified words are highlighted in bold.
  }  
  \resizebox{.48\textwidth}{!}{  \small
  \begin{tabular}{lp{6cm}}
    \toprule[1pt]
    (P) & How can I \textbf{lose weight} at age 55 ?\\
    (Q) & What are some ways to \textbf{lose weight} fast ?\\
    (P') & How can I \textbf{buy anything} at age 55 ?\\
    (Q') & What are some ways to \textbf{buy anything} fast ?\\
    Label & Positive\\
    Output & Positive $\rightarrow$ Negative\\
    \midrule
    (P) & If \textbf{infinite dark/vacuum/gravitational energy} can be created as universe expands , does it mean that their \textbf{potentiality} or potential \textbf{energy} is infinite ? \\
    (Q) & What are \textbf{good gifts} for a \textbf{foreign} visitor to bring when they 're invited to someone 's \textbf{home} in \textbf{Vietnam} for the first time ?\\
    (P') & If \textbf{local global interactions} can be created as universe expands , does it mean that their \textbf{existence} or potential \textbf{plane} is infinite ?\\
    (Q') & What are \textbf{global interactions} for a \textbf{local} visitor to bring when they 're invited to someone 's \textbf{plane} in \textbf{existence} for the first time ?\\
    Label & Negative\\
    Output & Negative $\rightarrow$ Positive\\
    \bottomrule[1pt]
  \end{tabular} }
  \label{case_bert_quora}
\end{table}

\begin{table}[!ht]
  \centering
  \caption{
  Typical modified examples for BERT as the target model on MRPC.
  }    
  \resizebox{.48\textwidth}{!}{  \small
  \begin{tabular}{lp{6cm}}
    \toprule[1pt]
    (P) & The \textbf{spacecraft} is scheduled to \textbf{blast} off as early as tomorrow or as late as Friday from the Jiuquan \textbf{launching site} in the \textbf{Gobi Desert} .\\
    (Q) & The \textbf{spacecraft} is scheduled to \textbf{blast} off between next Wednesday and Friday from a \textbf{launching site} in the \textbf{Gobi Desert} . \\
    (P') & The \textbf{match} is scheduled to \textbf{kick} off as early as tomorrow or as late as Friday from the Jiuquan \textbf{long day} in the \textbf{hot summer} . \\
    (Q') & The \textbf{match} is scheduled to \textbf{kick} off between next Wednesday and Friday from a \textbf{long day} in the \textbf{hot summer} . \\
    Label & Positive \\
    Output & Positive $\rightarrow$ Negative\\
    \hline
    (P) & The \textbf{resolution} was approved with no \textbf{debate} by delegates at the bar association 's annual meeting here .\\
    (Q) & Morales , who pleaded guilty in July , expressed `` sincere regret and \textbf{remorse} '' for his \textbf{crimes} .\\
    (P') & The \textbf{loss} was approved with no \textbf{surprise} by delegates at the bar association 's annual meeting here .\\
    (Q') & Morales , who pleaded guilty in July , expressed `` sincere regret and \textbf{surprise} '' for his \textbf{loss} .\\
    Label & Negative \\
    Output & Negative $\rightarrow$ Positive\\
    \bottomrule[1pt]
  \end{tabular}}
  \label{case_bert_mrpc}
\end{table}

We show some examples that our modification with shared words can make the target model fail, to further illustrate the robustness issue we reveal.
Table \ref{case_bert_quora} presents two examples using BERT as the target model on QQP.
For the first example (positive), changing from asking about ``lose weight'' to ``buy anything'' fools the target model to alter the predicted label, though the modified sentences are still asking about the same thing and are paraphrases.
For the second example (negative), introducing new shared words ``local'', ``global'', ``interactions'', ``existence'' and ``plane'' fools the target model to predict that the modified sentences are paraphrases, although the new sentences are still asking about different things.
Similarly, Table \ref{case_bert_mrpc} presents two examples on MRPC.